\documentclass[11pt]{article}

\makeatletter
\renewcommand{\maketitle}{
 \vspace*{0cm}  
  \begin{center}
    {\Large\bfseries \@title \par}
    \vspace{1em}
    
    {\large \@author \par}
    \vspace{1em}
    {\small \@date}
  \end{center}
  \@thanks
}
\makeatother

\usepackage[letterpaper, margin=1in]{geometry}
\usepackage{graphicx} 
\usepackage{natbib} 

\title{Language models as tools for investigating the distinction between possible and impossible natural languages\\[0.5em]
{\large Commentary on \citeauthor{futrell-mahowald-2025}'s BBS Article (\citeyear{futrell-mahowald-2025})
}}
\author{\textbf{Julie Kallini and Christopher Potts}\thanks{Equal contribution. Author order determined alphabetically by last name.}\\[0.2em]
Stanford University\\[0.2em]
\texttt{\{kallini,cgpotts\}@stanford.edu}
}
\date{December 5, 2025}

\begin{document}

\maketitle

\begin{abstract}\noindent
We argue that language models (LMs) have strong potential as investigative tools for probing the distinction between possible and impossible natural languages and thus uncovering the inductive biases that support human language learning. We outline a phased research program in which LM architectures are iteratively refined to better discriminate between possible and impossible languages, supporting linking hypotheses to human cognition.
\end{abstract}

Which conceivable linguistic systems are possible for humans to learn and use as natural languages? A complete answer to this question would yield profound insights into the human capacity for language. However, our tools for addressing the question are very limited. The historical record is shaped by external factors (e.g., patterns of migration and contact). Artificial language learning experiments happen in highly limited settings and cannot avoid influences from participants’ own languages, and any attempt to expose children only to impossible languages would be deeply unethical.

A number of recent papers present evidence that language models (LMs) learn possible human languages more efficiently than impossible ones. In light of how limited our existing toolkit is, this sounds like a promising development. Futrell and Mahowald interpret these studies as evidence that LMs possess inductive biases aligned with human languages, such as a preference for information locality.

What precisely is the value of such findings for the project of trying to understand what makes a language possible? On the face of it, the answer is not immediately clear. After all, LMs differ from humans in the data they consume, the data they produce, and how they learn. This might seem to make evidence derived from LMs irrelevant by definition.

Our own view (consistent with \citeauthor{futrell-mahowald-2025}'s perspective) is that LMs have great potential as investigative tools for understanding the possible/impossible distinction. We present our argument as a series of research phases:

\begin{description}
\item[Phase 1:] Find examples of possible and impossible languages. We might stipulate that any attested language is a possible language. For impossible languages, we assume there are clear cases that are very distant from the attested ones (e.g., languages that have no predictable structure). Linguists have also posited relatively clear instances of impossible languages that are closer to the attested ones \citep{moro-2008, moro-2023}, and these might be the most informative ones in terms of theory building.

\item[Phase 2:] Train LMs on minimal pairs of possible and impossible languages from Phase 1. \cite{mitchell-bowers-2020-priorless} helped begin this project. \cite{kallini-etal-2024-mission} present evidence that such models learn English more efficiently than counterfactual impossible languages. \cite{xu-etal-2025} and \cite{yang-etal-2025-anything} expand the empirical scope of these claims and arrive at similar conclusions. \cite{ziv-etal-2025} present a more mixed picture. Instances of alignment and misalignment are both valuable in this context.

\item[Phase 3:] Using the evidence gathered in Phase 2, study the inductive biases of these LMs and use these insights to inform statements about what the corresponding human biases are. Crucially, this will require us to state rigorous linking hypotheses between LM constructs and human cognition. The real challenge lies in stating linking hypotheses that are informative. Precedents from neuroscience make us optimistic about this project (e.g., \citealt{mcintosh-etal-2016-deep}).

\item[Phase 4:] Explore novel LM architectures, training datasets, and training objectives, seeking designs that are even more successful at distinguishing the possible from the impossible than those used in Phase 3. We see many opportunities. For example, many present-day LMs have mechanisms that give them what is, in effect, perfect memory over long sequences of words. By reducing the capacity of these mechanisms, we might encourage even more locality and thus better match human language. Implicit in this description is a hypothesis linking LM memory to human memory.
\end{description}

Once Phase 4 is reached, we return to Phase 2 for empirical evaluation against the languages we found in Phase 1. This leads to new linking hypotheses in Phase 3 and in turn to new LM innovations in Phase 4.

On the above approach, we do not immediately get insights into which languages are possible and which impossible (Phase 1). However, as we gain confidence in LMs as investigative tools, we might start to use their behavior with specific languages as evidence for the possible/impossible status for those languages. In this way, LMs could also help us expand our evidence base.

Importantly, on this approach, the LM acts as a tool for informing theories of language and cognition via linking hypotheses, and its value as a tool is measured by the power of those linking hypotheses. Various other tools could in principle play the role of the LM. For example, simple classifiers defined over the languages from Phase 1 could provide insights. LMs are privileged only insofar as they are the most successful language technologies ever created, and they invite many architectural modifications that could support viable linking hypotheses.

It seems likely to us that using LMs in this context would reopen some of the usual terms of the debate. For example, LMs do not inherently define a binary notion of ``learning a language''. To address this, we could (1) posit such a distinction for them, (2) change them in fundamental ways, or (3) change our conception of what it means to learn a language, to allow for more gradient notions of learning. Our own view is that human language learning is itself gradient and fluid, so we would likely opt for (3), but other researchers might respond differently. This is a question of how best to use LMs as tools here, and we think raising such questions is itself productive.

We are energized by the above project, and we already see evidence that it is yielding new insights. For example, as noted above, prior work suggests that the very general learning mechanisms used by today’s LMs suffice to create some observed linguistic locality effects. This is in itself illuminating about the factors that can lead to these locality effects. What sort of modifications to those mechanisms would increase this alignment? In addition, \cite{hunter-2025} argues for the importance of linguistic constituent structure in any discussion of locality effects, and he describes new candidate possible/impossible language pairs designed to engage with this issue. What class of LMs is able to capture this asymmetry, and what might such LMs tell us about language and cognition?

\bibliographystyle{apalike}
\bibliography{bibliography}

\end{document}